\documentclass{article}
\usepackage[preprint]{corl_2024} 
\usepackage{wrapfig}
\usepackage{graphicx}
\usepackage{tikz}
\usepackage{caption}
\usepackage{csquotes}
\usepackage{subcaption}
\usepackage[utf8]{inputenc}
\usepackage{pgfplots}
\DeclareUnicodeCharacter{2212}{−}
\usepgfplotslibrary{groupplots,dateplot}
\usetikzlibrary{patterns,shapes.arrows}
\pgfplotsset{compat=newest}
\usepackage{printlen} 
\usepackage{booktabs}
\usepackage{tabularx}
\usepackage{listings}
\usepackage{multirow}

\definecolor{codegreen}{rgb}{0,0.6,0}
\definecolor{codegray}{rgb}{0.5,0.5,0.5}
\definecolor{codepurple}{rgb}{0.58,0,0.82}
\definecolor{backcolour}{rgb}{0.95,0.95,0.92}

\title{CogExplore: Contextual Exploration with Language-Encoded Environment Representations}

\author{
  Harel Biggie $^*$\\
  Department of Computer Science\\
  University of Colorado Boulder
  United States\\
  \texttt{harel.biggie@colorado.edu} \\
  \And
  Patrick Cooper $^*$ \\
  University of Colorado Boulder \\
  \texttt{patrick.cooper@colorado.edu} \\
  \AND
   Doncey Albin \\
  University of Colorado Boulder \\
   \texttt{doncey.albin@colorado.edu} \\
   \And
    Kristen Such \\
  University of Colorado Boulder \\
   \texttt{kristen.such@colorado.edu} \\
   \And
   Christoffer Heckman \\
   University of Colorado Boulder \\
  \texttt{christoffer.heckman@colorado.edu} \\
}

\begin{document}
\maketitle

\def\thefootnote{*}\footnotetext{These authors contributed equally to this work.}


\begin{abstract}

    Integrating language models into robotic exploration frameworks improves performance in unmapped environments by providing the ability to reason over semantic groundings, contextual cues, and temporal states. The proposed method employs large language models (GPT-3.5 and Claude Haiku) to reason over these cues and express that reasoning in terms of natural language, which can be used to inform future states. We are motivated by the context of search-and-rescue applications where efficient exploration is critical. We find that by leveraging natural language, semantics, and tracking temporal states, the proposed method greatly reduces exploration path distance and further exposes the need for environment-dependent heuristics. Moreover, the method is highly robust to a variety of environments and noisy vision detections, as shown with a 100\% success rate in a series of comprehensive experiments across three different environments conducted in a custom simulation pipeline operating in Unreal Engine.
\end{abstract}



\section{Introduction}
Exploring unmapped environments is paramount to modern robotic applications, such as search-and-rescue and assistive robotics.  While state-of-the-art methods have made incredible progress in exploring in challenging search and rescue scenarios \cite{biggie2023flexible, tranzatto2022cerberus, agha2021nebula}, they typically rely on hand-engineered heuristics based on the geometry of the environment, e.g. optimizing for volumetric gain. Such hand-engineered features fail to incorporate the rich semantics and contextual cues of an environment \cite{biggie2023tell}.

\begin{figure}[!htpb]
  \centering
  \includegraphics[width=0.8\textwidth]{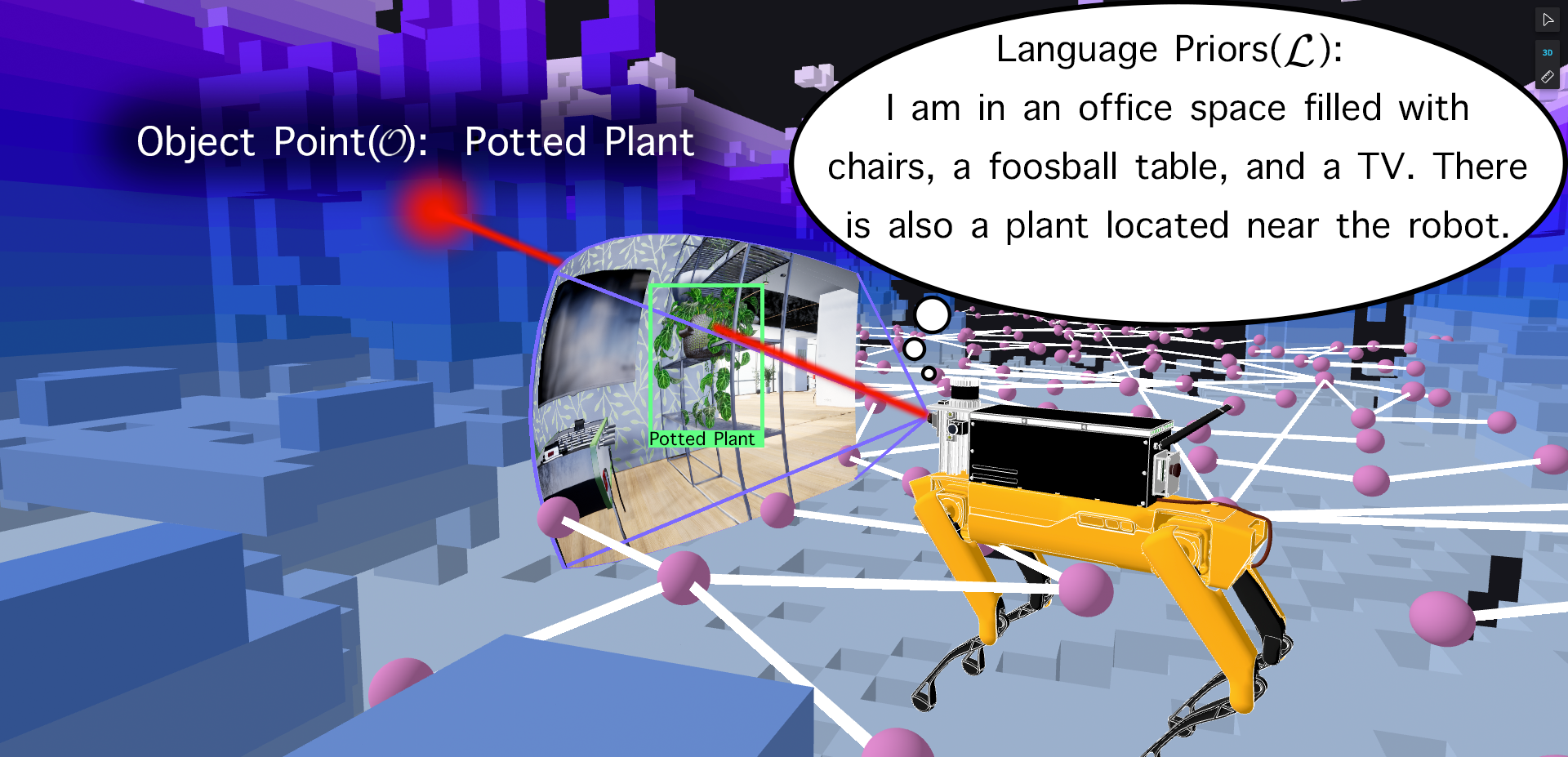} 
  \caption{Spot characterizing its environment through its VQA model (Language Priors), searching for specific objects with its object detection model and creating projections (Object Points shown in red) surrounded by a set of  navigable graph points (shown in purple).} 
  \label{fig:intro}
\end{figure}

Given the problem of exploring a previously unknown environment, we develop a framework, Contextual Exploration (Cog Explore), that leverages foundation models, or large language models (LLMs) with natural-language-based representations of the environment to perform navigation using both geometric and contextually rich features, as well as temporal states.

To effectively leverage natural-language, we must ground language utterances to the physical world in which the robot operates. Various approaches have been used to associate language with the physical domain, ranging from probabilistic graph-based structures \cite{tellex2011understanding,howard2014natural,kollar2010toward} to end-to-end learning-based methods \cite{driess2023palm}. We represent the grounding problem as a probabilistic set of planning points, object points, and language priors obtained from a set of vision models. Our framework then selects the next best location to explore given this set of priors and a goal. Specifically, we leverage foundation models \cite{DBLP:journals/corr/abs-2005-14165} \cite{jiang2024mixtral} which have shown the ability to exhibit remarkable contextualization and reasoning by utilizing efficient next token predication.

Navigation solutions powered by LLMs offer distinct advantages over metric based methods such as frontier point finding \cite{yamauchi1997frontier}. For instance, a robot operating in a school instructed to ``Go find a whiteboard'' would spend equal time exploring a bathroom or a classroom when using a frontier based planner. However, by integrating an LLM's world knowledge into search strategies, the resulting planning method can leverage the fact that erasers are for more likely to be found in a classroom. Moreover, we can rely on other cues in the environment to influence the search behavior. For example, if a whiteboard is present, the robot should be more exploitative than explorative. This adaptive form of reasoning, aiming for the most plausible explanation—or abductive reasoning allows CogExplore to create a form of working memory.

LLMs have shown remarkable capabilities in various domains, yet they do not explicitly encode a time dimension, meaning their appreciation of time is implicit \cite{deng2024facing}. Specifically, LLMs encode memories to be used in future reasoning implicitly as an autoregressive relation instead of an explicitly modeled recurrent relation \cite{DBLP:journals/corr/VaswaniSPUJGKP17}. We introduce a method to simulate working memory by compressing prior states and justifications into a concise log. By leveraging temporal and contextual reasoning, CogExplore guides robots to navigate diverse environments, while adapting the exploration strategy as new information is acquired. 

Our method presents possibilities for advances in scenarios where environments are semantically rich, but of which very little is known a priori.  Such environments included domestic interiors, where robots might be asked to help locate and retrieve household items (\textit{Bring the groceries in from the garage}), identify conditions (\textit{Check if the oven is on}), or interact with humans (\textit{Open the front door and guide the people to my room}). We validate CogExplore's exploration capabilities with 210 simulations of 45 minutes each operating in 3 different environments across 7 different tasks and show the method is capable of performing temporal, geometric, and semantic reasoning.






\vspace{-12pt}

\section{Related Works}

\label{sec:citations}

\subsection{Robotic Exploration}
Graph-based approaches have proven effective in robotic exploration \cite{tellex2011understanding,howard2014natural,kollar2010toward, lavalle1998rapidly, noreen2016optimal, karaman2011anytime, lan2015continuous, yamauchi1997frontier}. These methods construct topological representations of the environment to guide exploration and planning. Recent works have incorporated semantics into these graph-based frameworks to enhance performance \cite{crespo2020semantic}.

In addition to graph-based techniques, neural networks have been applied to robotic navigation, often serving as heuristics \cite{mac2016heuristic}. Leveraging advancements in computer vision, modern perception models have been integrated with traditional exploration approaches, leading to improved capabilities \cite{gadre2022clip, huang2022visual}.

\subsection{LLM-Powered Robotic Navigation}
LLMs have emerged as powerful tools for robotic navigation and manipulation. By encoding scene dynamics and task specifications in natural language, LLMs enable robots to reason about their environment and objectives at a high level \cite{zeng2023large, ahn2022can}. 

LLMs can act as agents capable of decomposing and executing complex tasks through modular prompting \cite{liu2023pre, gupta2023visual, huang2022language}. They have demonstrated proficiency in code generation for robotic applications \cite{madaan2022language, vaithilingam2022expectation, nijkamp2022codegen, saycan2022progprompt}. The combination of embodied agents for low-level skills and LLMs for high-level reasoning has shown promise \cite{brohan2023can}.

End-to-end neural network approaches such as LATTE \cite{bucker2022latte} and CLIPort \cite{shridhar2021cliport} map natural language intentions to robot actions. HULC \cite{mees2022hulc} combines language conditioning and semantic knowledge for efficient imitation learning. While this approach works well in specific domains, it is not able to learn from the voluminous unsupervised datasets ubiquitous to LLMs.

\begin{figure}[h!]
    \centering
        \begin{subfigure}[b]{0.3\textwidth}
        \centering
        \includegraphics[width=\textwidth]{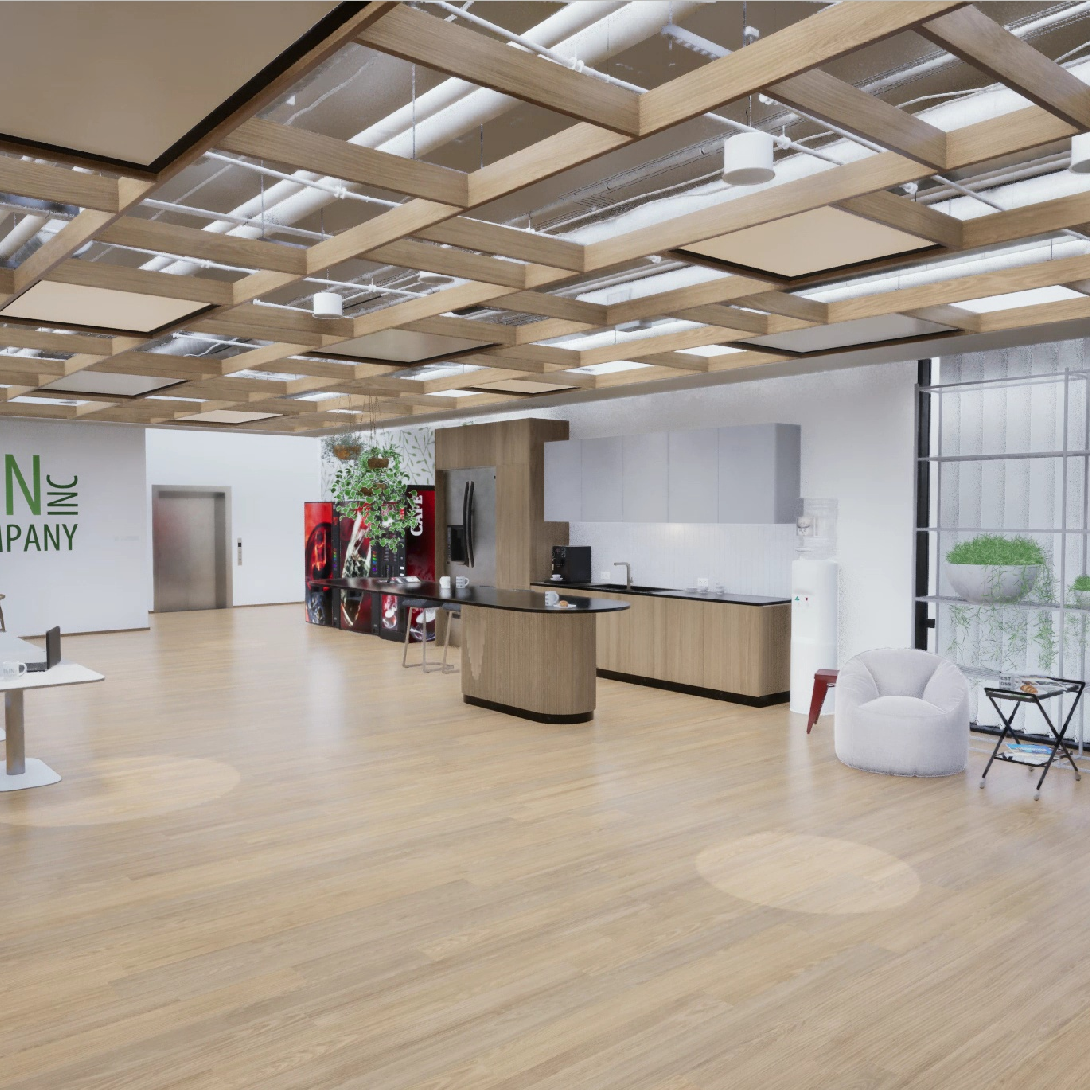}
        \caption{Office 1}
        \label{fig:office1ex}
    \end{subfigure}
    \hfill 
    \begin{subfigure}[b]{0.3\textwidth}
        \centering
        \includegraphics[width=\textwidth]{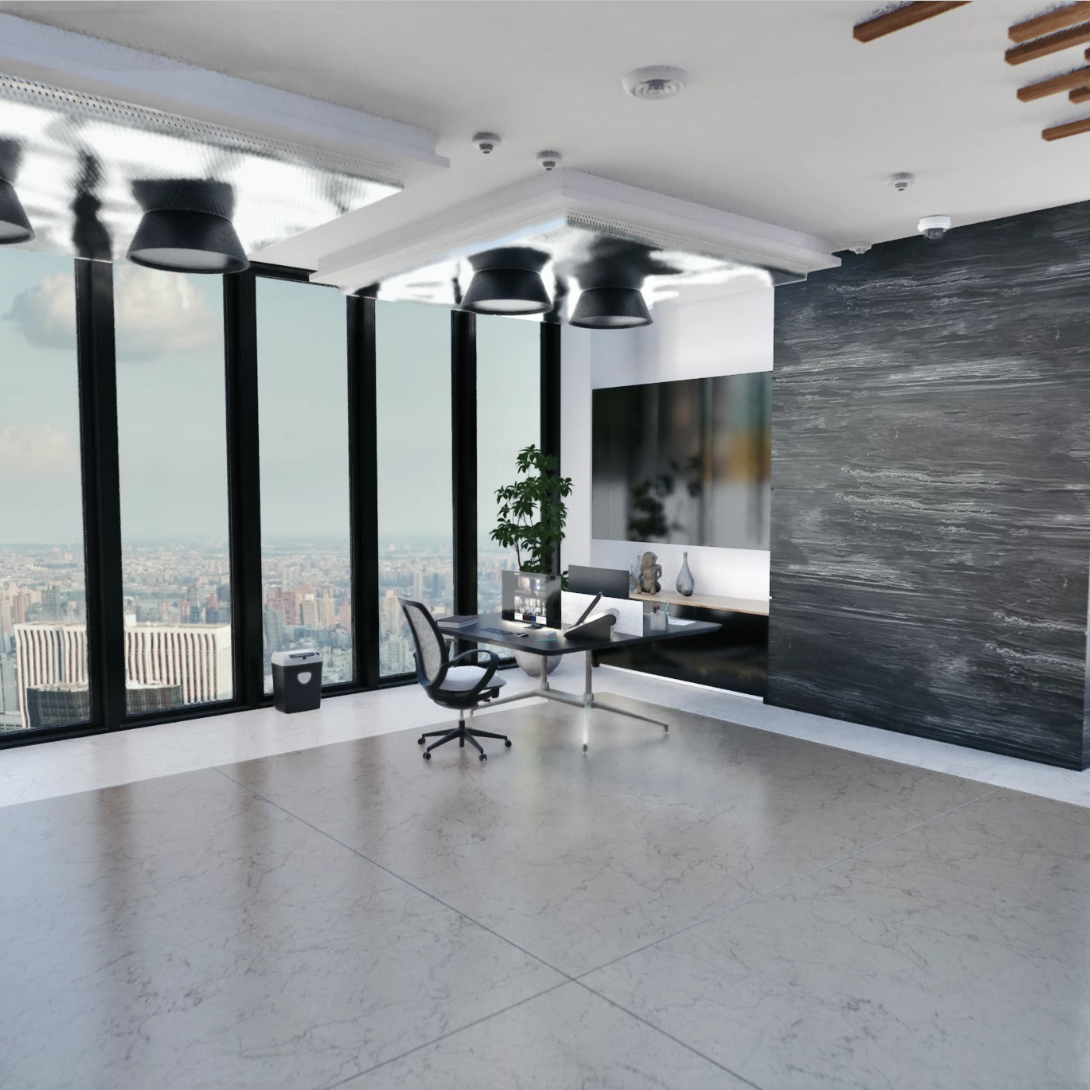}
        \caption{Office 2}
        \label{fig:office2ex}
    \end{subfigure}
    \hfill
    \begin{subfigure}[b]{0.3\textwidth}
        \centering
        \includegraphics[width=\textwidth]{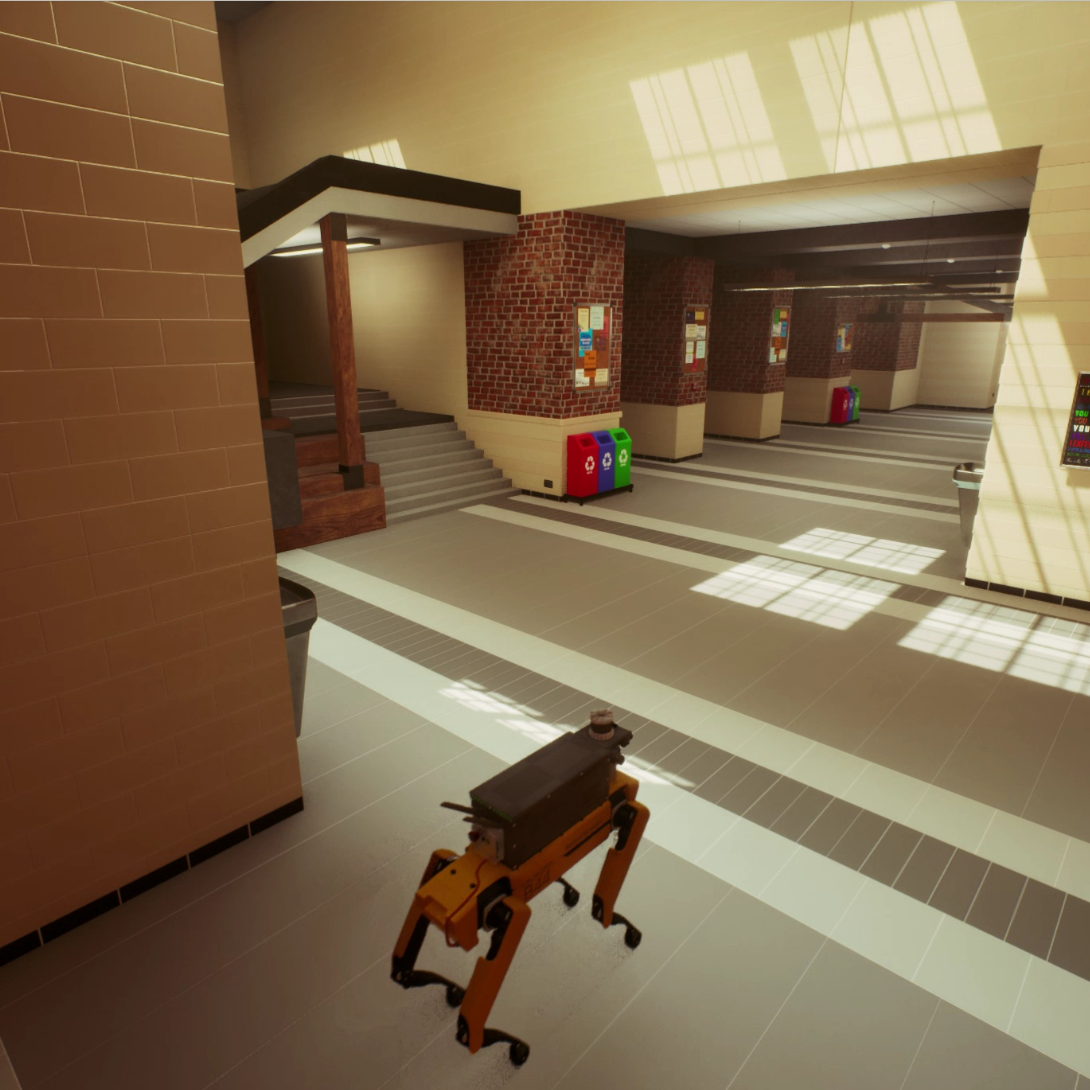}
        \caption{School}
        \label{fig:schoolex}
    \end{subfigure}
    \hfill
    \caption{Renderings from Unreal Engine Environments}
    \label{fig:unrealex}
\end{figure}

Grounding natural language instructions to the physical environment is crucial for LLM-powered robotic navigation. Neuro-symbolic approaches \cite{mao2019neuro} and learned traversability functions \cite{shah2021ving} have been explored for this purpose. PromptCraft \cite{vemprala2023chatgpt} tackles the challenge of describing complex navigation tasks through prompt engineering. ORION \cite{dai2023think} demonstrates personalized, interactive object navigation using natural language.

While LLMs excel at reasoning over unstructured data \cite{floridi2020gpt, touvron2023llama, tellex2020robots, zhao2023survey} using transformer architectures \cite{vaswani2017attention}, guiding them to desired outcomes and maintaining temporal coherence remains an open problem. Recent works integrating LLMs with object scene representation transformers for task and motion planning \cite{brohan2023can, driess2023palm, saycan2022progprompt} still lack the interpretability and guarantees of traditional methods \cite{noreen2016optimal}. 

The semantic reasoning abilities of LLMs have been leveraged to guide exploration and planning in novel environments \cite{shah2023navigation, shah2021lmnavbench}. LLMs have also been utilized for semantic grounding in complex outdoor environments using contextually relevant instructions \cite{shah2023lm}. Our method introduces new capabilities to this body of work, allowing LLMs to selectively encode relevant states based on environment descriptions and then reason directly over these natural language encoded states, along with temporal states derived from a log of past movements.


\section{CogExplore}
\label{sec:methodology}

\begin{figure}
    \centering
    \includegraphics[width=0.99\linewidth]{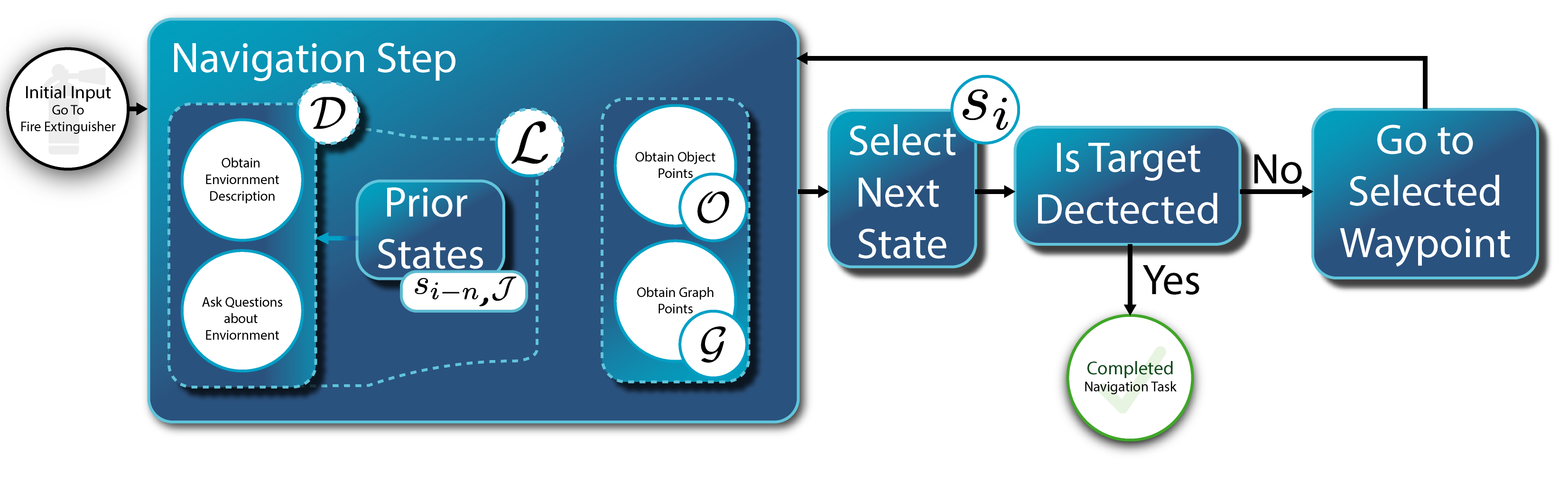}    
    \caption{CogExplore System Diagarm}
    \label{fig:pipeline}
\end{figure}

Our proposed exploration framework, CogExplore, represents relevant features of the environment as natural language strings. Formally, we represent possible states for the robot as, $\mathcal{S} = \{s_1, s_2, ..., s_n\}$ where  $s_i$ is a possible state for the robot. At each state the robot has a set of planning graph points $\mathcal{G} = \{g_1, g_2, ..., g_k\}$, a set of object points  $\mathcal{O} = \{o_1, o_2, ..., o_m\}$ and a set of language priors $\mathcal{L}$. Graph points represent traversable areas of the environment, and we represent them as a list in text from $\mathcal{G}_i = \{x,y,z\}$. Object points are represented similarly but with a corresponding label and probability ($\mathcal{O}_i = \{x, y, z, c, p\}$ where $c$ is the class and $p$ is the confidence represented as a probability), obtained from an open vocabulary object detector. Specifically, we utilize YoloWorld \cite{Cheng2024YOLOWorld} and Segment Anything \cite{kirillov2023segment} to project open vocabulary detections into 3D points. Examples of $\mathcal{G}, \mathcal{L}$, and $\mathcal{O}$ can be seen in Figure \ref{fig:intro} and full details of, the open vocabulary 3D detection system can be found in Section \ref{appendix:openvocab} of the Appendix. The language priors, $\mathcal{L}$, consist of three components; environmental descriptions, prior robot states, and the justification for choosing the next state. Environment descriptions $\mathcal{D}$ are obtained from a series of questions generated by foundation model and answered using a multimodal visual question and answering model (VQA) called LLaVA -1.5 \cite{liu2024llavanext}, as shown in Figure \ref{fig:pipeline}. Prior states ($s_{i-n}$) and the justification for choosing the state ($\mathcal{J}_{i-n}$) and the full set of language priors is $\mathcal{L} = \{\mathcal{J}_{i-n}, s_{i-n}, \mathcal{D}\}$ along with the model's justification for selecting the state. In order to select the next best state, we can model the task as a maximum likelihood estimate:
\begin{equation}
    \arg\max_{s_1, s_2, ..., s_t} P(s_g | s_t, \mathcal{S}, \mathcal{O}, \mathcal{G}, \mathcal{L}) \prod_{i=1}^t P(s_i | s_{i-1}, \mathcal{S}, \mathcal{O}, \mathcal{G}, \mathcal{L})
\end{equation}

At each iteration, the foundation model is asked to direct the robot to the state that is most likely to find the goal state, $s_g$ given the robot's past states and any new observations. The process repeats until the robot arrives at $s_g$.

\subsection{Exploration Planner}
CogExplore's autonomous navigation uses a state-of-the-art graph based exploration planner with frontier point finding \cite{2020dangGBPlanner} to help the language model select reliable states. Realtime mapping is performed using the a voxel based map \cite{hornung2013octomap}. The planner was one of the top performing exploration method \cite{tranzatto2022team} at the DARPA SubT \cite{DARPA2022} challenge aimed at advancing exploration capabilities. At each planning iteration, a set of sparse global graph points and dense local points are generated. We employ a 2D Gaussian function to sparsity the distribution of these points, where weights are assigned based on their proximity to the robot's current location. The result is a distribution of low density points in distant areas with a high concentration of points near the agent allowing CogExplore to plan with high granularity locally while still having the ability to explore regions of interest that are further away.  The total sets of points is encoded into $\mathcal{G}$.

\subsection{Foundation Model Prompting Schemes}

We rely on language foundation models to perform four tasks: Generating questions about the environment, generating object labels, compressing state information, and performing state selection. Foundation models are highly sensitive to the particular presentation of the underlying information \cite{chen2023models}. To assist the model's geometric reasoning, we label some of the graph points ($\mathcal{G}$) as new, if they are in an unexplored region. A point is labeled as new if it is beyond a certain distance threshold from all existing points in the graph. This distance threshold is determined using a KD-Tree \cite{bentley1975multidimensional} structure to efficiently compute the nearest neighbors.

All of our prompts are highly modular in construction allowing us to incorporate custom visual grounding information ($\mathcal{O}, \mathcal{L})$, prior states $s_{i-n}$, navigation points ($\mathcal{G}$), and specific instructions designed to handle the strengths and weaknesses of a particular model. In practice, this accommodated the willingness of GPT to select distant points with a high potential for novelty relative to the systematic strategy adopted by Claude. We used the same prompts across all experiments, and the full prompt details can be found in Section \ref{appendix:prompting} of the Appendix.

For each iteration of the exploration cycle, the LLM is given a form to fill out, which creates a set of fields for the next waypoint output, as well as fields for a characterization of the environment and a justification field to insert reasoning for the selection of the particular waypoint it chose. All of our prompts follow this mad lib style form:

\begin{lstlisting}
I need to select a waypoint from a numbered list of graph points, frontier points and object points. The point I have selected is point number: [insert the graph, frontier or object point numeral number here]. I am selecting this point because I believe it makes strategic sense to get me closer to solving my navigation task "*INSERT_QUERY_HERE*". My environment can be described as [describe the environment]. My reasoning is that this point [insert your reasoning here].
\end{lstlisting}

Once the generation is complete, the LLM is queried again to compress the prior output into a concise state of 50-100 words ($\mathcal{J}$) these compressed states with justifications enable CogExplore to reason over its past states, facilitating a contiguous exploration process. Each of these logs is appended to a memory window with some fixed length, in practice we discovered a length of 10 was appropriate for both GPT-3.5-Turbo and Claude Haiku.

\section{Experimental Results}

\subsection{Experimental Setup}

We evaluate the performance of CogExplore's ability explore and find a target object using a simulated Boston Dynamics Spot in three different environments using Unreal Engine \cite{unrealengine}. We compare CogExplore against a $A^*$ direct path metric and the Vision-Enabled Frontier Exploration Planner (VEFEP) described below.  

Two office scenes (Office 1, Office 2) and a more complex school environment (School) were designed, which provided the structure for a total of 7 object-retrieval scenarios to be tested, as shown in Table \ref{tab:env}. In Office 1, there were three tasks involving locating a fire extinguisher in different locations, denoted as FE1, FE2, and FE3. Office 2 featured two tasks: finding a coffee table ($CT$) and an office chair ($OC$). In School, the tasks were to locate a whiteboard ($WB$) and a bookshelf ($BS$).


For each task, we simulate the navigation capabilities of a Boston Dynamics Spot and allow the robot to explore for 45 minutes. A run terminates once the target object is found. We conduct 15 trials per task on Azure ``NC64as\_T4\_v3'' instances with 4 Nvidia T4 GPUs. We utilize 3 GPUs per simulation, one for the unreal simulator, one for the object detector, and one for the VQA model. We note that for Office 1, and the school scene, the simulator runs in real-time. While in Office 2 the simulator runs in half-time due to higher quality graphical assets. While CogExplore does run in real-time, with the assistance of cloud computing we find runtimes are not an accurate measurement of exploration performance due to uncontrollable factors such latency and API call throttling from leading foundation model providers.

\begin{table}[h!]
\footnotesize
\resizebox{\textwidth}{!}{
\begin{tabular}{|lll|lll|lll|}
\hline
\multicolumn{3}{|c|}{\textbf{Office 1 (572 $m^2$):  Figure \ref{fig:office1ex}}}                                                                                                                                                                                          & \multicolumn{3}{c|}{\textbf{Office 2 (1450 $m^2$): Figure \ref{fig:office2ex}}}                                                                                                                                                                                                   & \multicolumn{3}{c|}{\textbf{School (1287 $m^2$): Figure \ref{fig:schoolex}}}                                                                                                                                                                                          \\ \hline
\multicolumn{1}{|l|}{\textbf{Task}}                                                        & \multicolumn{1}{l|}{\textbf{\begin{tabular}[c]{@{}l@{}}Direct \\ Path (m)\end{tabular}}} & \textbf{\begin{tabular}[c]{@{}l@{}}VEFEP \#\\ Timeout\end{tabular}} & \multicolumn{1}{l|}{\textbf{Task}}                                                                 & \multicolumn{1}{l|}{\textbf{\begin{tabular}[c]{@{}l@{}}Direct \\ Path (m)\end{tabular}}} & \textbf{\begin{tabular}[c]{@{}l@{}}VEFEP \#\\  Timeout\end{tabular}} & \multicolumn{1}{l|}{\textbf{Task}}                                                               & \multicolumn{1}{l|}{\textbf{\begin{tabular}[c]{@{}l@{}}Direct \\ Path (m)\end{tabular}}} & \textbf{\begin{tabular}[c]{@{}l@{}}VEFEP \#\\  Timeout\end{tabular}} \\ \hline
\multicolumn{1}{|l|}{\begin{tabular}[c]{@{}l@{}}Fire Extinguisher \\ 1 (FE1)\end{tabular}} & \multicolumn{1}{l|}{25.2}                                                                & 0                                                                   & \multicolumn{1}{l|}{\begin{tabular}[c]{@{}l@{}}Office \\ Chair (OC)\end{tabular}}                  & \multicolumn{1}{l|}{33.7}                                                                & 3                                                                    & \multicolumn{1}{l|}{\begin{tabular}[c]{@{}l@{}}Whiteboard\\ Eraser \\ (WE)\end{tabular}}         & \multicolumn{1}{l|}{26.1}                                                                & 5                                                           \\ \hline
\multicolumn{1}{|l|}{\begin{tabular}[c]{@{}l@{}}Fire Extinguisher \\ 2 (FE2)\end{tabular}} & \multicolumn{1}{l|}{27.1}                                                                & 0                                                                   & \multicolumn{1}{l|}{\multirow{2}{*}{\begin{tabular}[c]{@{}l@{}}Coffee \\ Table (CT)\end{tabular}}} & \multicolumn{1}{l|}{\multirow{2}{*}{26.2}}                                               & \multirow{2}{*}{6}                                                   & \multicolumn{1}{l|}{\multirow{2}{*}{\begin{tabular}[c]{@{}l@{}}Book-\\ shelf (BS)\end{tabular}}} & \multicolumn{1}{l|}{\multirow{2}{*}{13}}                                                 & \multirow{2}{*}{2}                                          \\ \cline{1-3}
\multicolumn{1}{|l|}{\begin{tabular}[c]{@{}l@{}}Fire Extinguisher \\ 3 (FE3)\end{tabular}} & \multicolumn{1}{l|}{15.6}                                                                & 0                                                                   & \multicolumn{1}{l|}{}                                                                              & \multicolumn{1}{l|}{}                                                                    &                                                                      & \multicolumn{1}{l|}{}                                                                            & \multicolumn{1}{l|}{}                                                                    &                                                             \\ \hline
\end{tabular}
}
\caption{Simulation Environments and Corresponding Tasks. Each run was given 45 minutes of simulation time to complete. None of the CogExplore runs timed out, and the VEFEP timeouts are shown here.}
\label{tab:env}
\end{table}

We compare the performance of CogExplore running with Anthropic's \emph{claude-3-haiku-20240307} (CE-H) and OpenAI'S \emph{gpt-3.5-turbo-0125} (CE-3.5) as the backend models to a baseline Vision-Enabled Frontier Exploration Planner (VEFEP). VEFEP calculates the exploration gain heuristic \( g \) for a given path \( \sigma_{i} \) by summing the volumetric gain \(\mathbf{VG}\) weighted by a function for each vertex in the path \cite{Bachrach2013TrajectoryBE}. Parameters are based on the ones used by the authors in \cite{2020dangGBPlanner}. This planner explores until it detects the target object using the same 3D open Vocabulary detector used by CogExplore and described in section \ref{sec:methodology}. All constructed environments, simulation code, and CogExplore code will be released upon publication.

\begin{figure}[htbp]
  \centering
  \includegraphics[width=\textwidth]{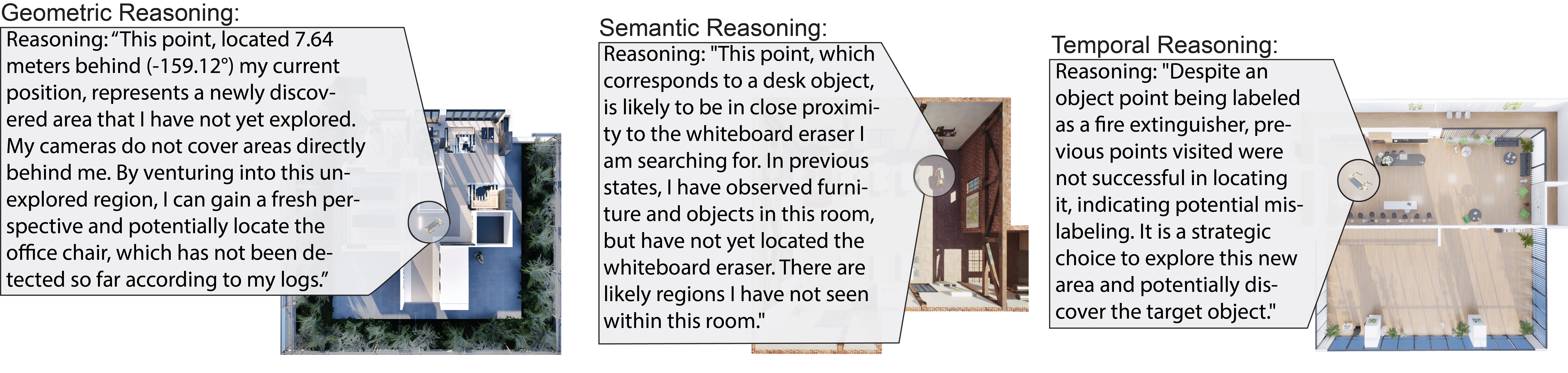} 
  \caption{Example Runs Demonstrating Varieties of Reasoning}
  \label{fig:exampleruns}
\end{figure}

\begin{figure}[h!]
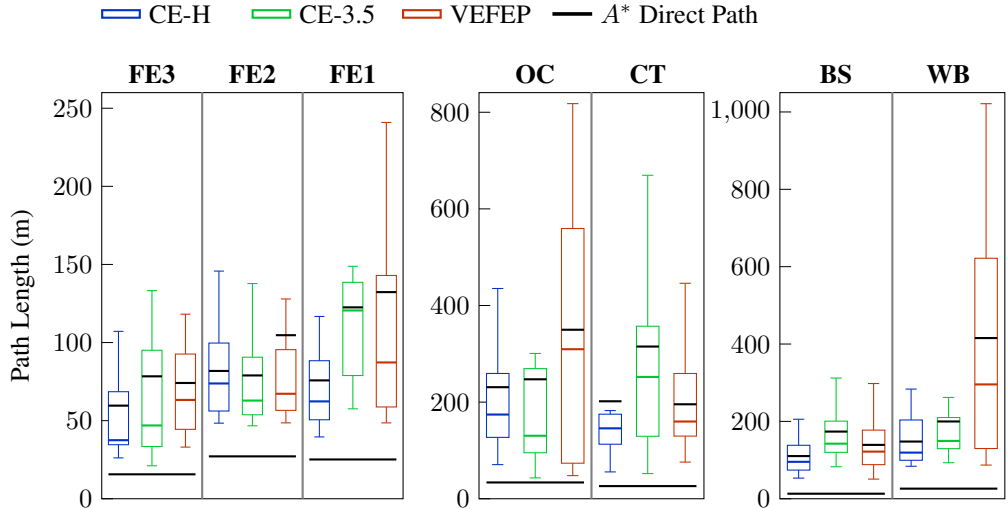

    \centering
\begin{tikzpicture}[rotate=90]

\definecolor{claude}{RGB}{0,50, 200}
\definecolor{vefep}{RGB}{200, 50, 0}
\definecolor{gpt}{RGB}{0, 200, 50}

\definecolor{darkgray176}{RGB}{176,176,176}
\definecolor{gray}{RGB}{128,128,128}
\definecolor{lightcyan227249251}{RGB}{227,249,251}
\definecolor{lightgray204}{RGB}{204,204,204}
\definecolor{lightskyblue156168253}{RGB}{156,168,253}
\definecolor{midnightblue2626102}{RGB}{26,26,102}
\definecolor{slateblue102102204}{RGB}{102,102,204}
\definecolor{slateblue7777178}{RGB}{77,77,178}
\definecolor{lavender206217248}{RGB}{206,217,248}

\draw[thick, claude, fill=white] (6.5, 12) rectangle (6.4, 11.5);
\node[right] at (6.45, 11.5) {CE-H};

\draw[thick, gpt, fill=white] (6.5, 10) rectangle (6.4, 9.5);
\node[right] at (6.45, 9.5) {CE-3.5};

\draw[thick, vefep, fill=white] (6.5, 8) rectangle (6.4, 7.5);
\node[right] at (6.45, 7.5) {VEFEP};

\draw[thick, black, fill=black] (6.46, 6) rectangle (6.44, 5.5);
\node[right] at (6.45, 5.5) {$A^*$ Direct Path};

\input{figures/tikz/OfficeEnv1_boxplots}
\input{figures/tikz/OfficeEnv2_boxplots}
\input{figures/tikz/Classroom01_boxplots}

\end{tikzpicture}
    \caption{Path length comparisons for each method (CE-3.5, CE-H, VEFEP) on completing each of the seven tasks. Black line for each whisker plot is the mean.}
    \label{fig:boxplots}
\end{figure}

\textbf{Quantitative Results:} From Figure \ref{fig:boxplots} we can see that in Office 1, the smallest environment, VEFEP, CE-H, and CE-3.5 all have similar mean path lengths. As the environment size increases in Office 2, making the exploration tasks more complex, we observe that on the office chair task both CE-H and CE-3.5 outperform VEFEP. We also note that VEFEP, timed out in 3 out of 15 of the runs as noted in Table \ref{tab:env}. VEFEP had the lowest median run length in the coffee table task. However, it also failed to find the table in 6 out of 15 experiments, as shown in Table \ref{tab:env}. A similar result is present in the school environment with the bookshelf task where CE-H outperforms VEFEP on path length but CE-3.5 does not. However, both CE methods complete the task 100 percent of the time and VEFEP timed out in 2 out of 15 tasks. In the whiteboard eraser task, both CE models significantly outperform VEFEP, which also had 5 out of 15 failures or a 33.3\% failure rate. Of note, is that \emph{none} of the CE methods failed at the exploration task in any of the environments. 

\textbf{Qualitative Results:} In Figure \ref{fig:birdseyeviews} we observe that in the Whiteboard Eraser task CE-H and CE-3.5 only enter the room with the eraser once whereas VEFEP enters, exits and proceeds to loop around the environment. Similarly, in the Go to the Office Chair task CE-H and CE-3.5, immediately go to the chair and take more efficient paths to get there. In contrast, VEFEP heads outside and explores the areas significantly before returning to indoors to find the chair.

We provide anecdotal examples of the model's reasoning in Figure \ref{fig:exampleruns}, where we see examples of the model performing spatial reasoning (left), semantic reasoning (center) and temporal reasoning (right). At each iteration we can probe why the model made a decision which highlights the explainability of the CogExplore method.




\section{Discussion}
Our results highlight the robustness of CogExplore's exploration abilities with a \emph{100\% success rate} across all environments with both GPT and Haiku variants, in contrast to VEFEP's 15.2\% overall failure and  40\% failure rate at the coffee table task. Moreover, CogExplore's use of natural language justifications for each state selection allow us to directly probe why the framework is acting in a certain manner and understand the rationale behind the more efficient exploration paths. From these justifications we can see that he model is capable of performing geometric reasoning, semantic reasoning, temporal reasoning and fault-tolerant reasoning.

Cog Explore's ability to leverage a variety of reasoning mechanisms through encoding environmental features into natural language strings enables the sharp performance gains from heuristic based methods like VEFEP. Frontier-based exploration methods like VEFEP rely on hand tuning of the volumetric gain weights in order to influence the exploration behavior, whereas CogExplore adapts based on observations. This is highlighted in the example where our framework tells the robot to choose a further state because no clues were found indicating the fire extinguisher is nearby:

\begin{lstlisting}
"The point is far enough to explore a different vantage point within the domestic setting as no indications of the fire extinguisher are present."
\end{lstlisting}

\textbf{Geometric Reasoning.} We see geometric reasoning on the right side of Figure \ref{fig:exampleruns} where the model explicitly chooses a point behind the robot. It is aware that the robot can not detect objects, directly behind it, since there is no rear camera. This variety of intuitive geometric reasoning directly contributes to the shorter exploration paths lengths demonstrated by CogExplore.

\begin{figure}[h!]
    \centering
    \begin{subfigure}{0.48\textwidth}
        \includegraphics[width=\textwidth]{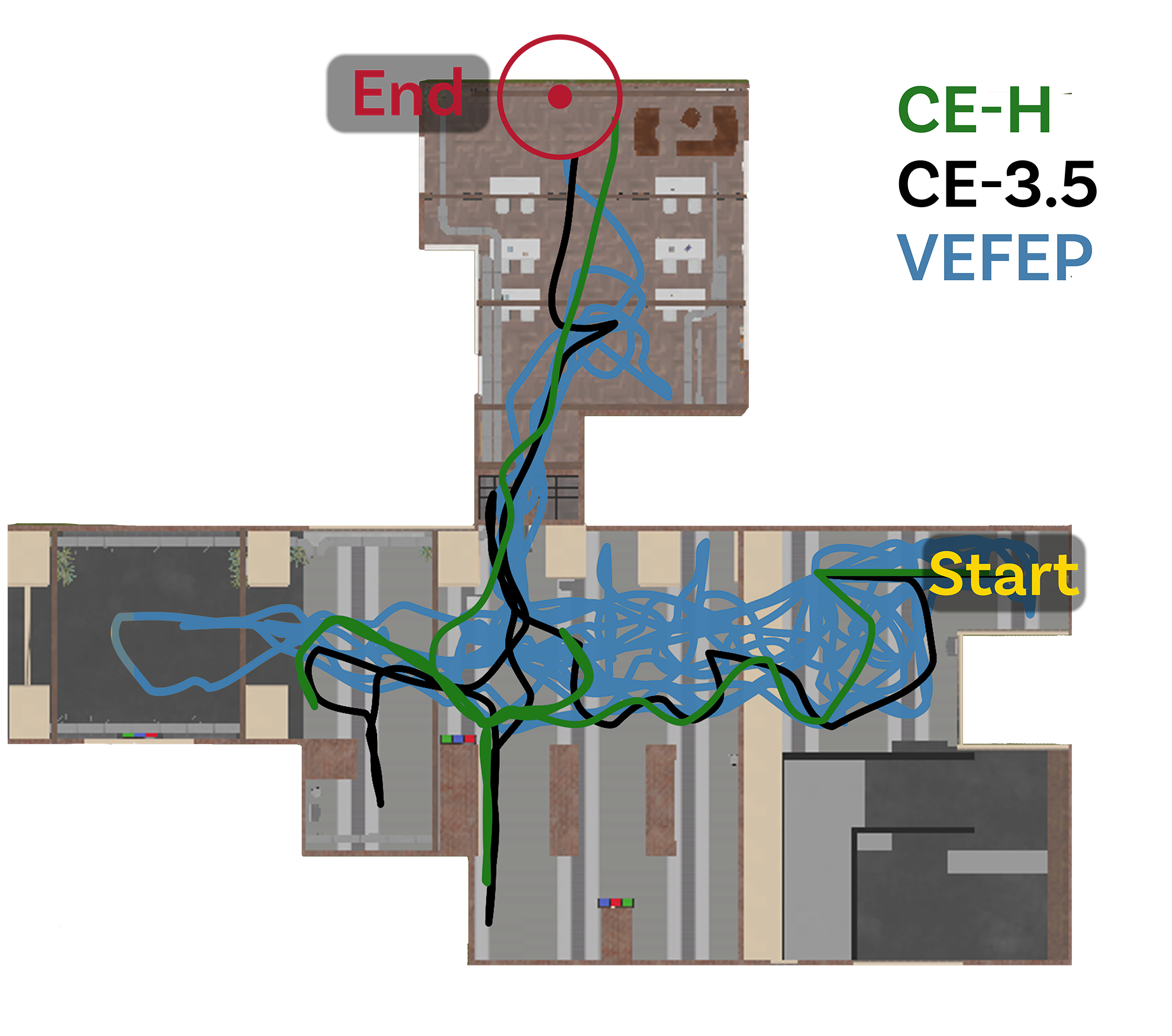}
        \caption{Note that the VEFEP  re-visits the same places several times before eventually locating the goal object, whereas the CogExplore is able to more effectively reason over unexplored regions.}
        \label{fig:eraser_path}
    \end{subfigure}
    \hfill
    \begin{subfigure}{0.48\textwidth}
        \includegraphics[width=\textwidth]{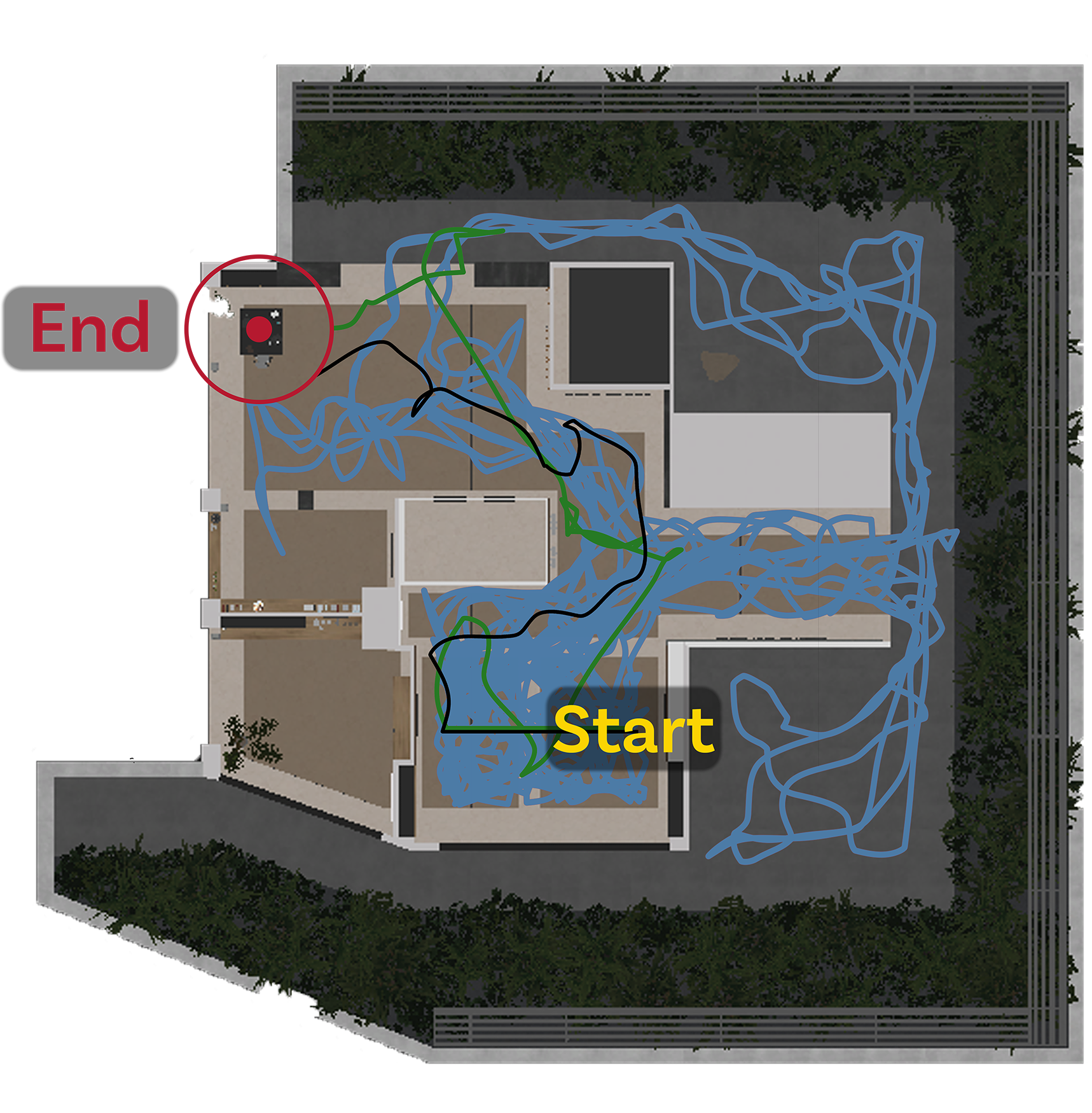}
        \caption{Note that VEFEP obtains noisy projections and leaves the area rather than continuing to search the area for the office chair. Whereas Cog Explore is able to reason over the noisy projections.}
        \label{fig:office_chair}
    \end{subfigure}
    \caption{Bird's Eye Views: Go to the Whiteboard Eraser, School (a) and Go to the Office Chair, Office 2 (b) }
    \label{fig:birdseyeviews}
\end{figure}

\textbf{Semantic Reasoning.} Another key factor that contributes to robust and efficient performance is the model's ability to semantically reason over cues in the environment. In the center of Figure \ref{fig:exampleruns} we see the model choosing a nearby point because a desk was found which could contain a white board eraser (the desired object in the exploration task). Both objects are common located in classroom and office settings, hence their share semantic similarity. Traditional frontier based planners continue exploring other areas based on the original heuristic for volumetric gain despite these cues, which is also evident by the path taken by VEFEP in Figure \ref{fig:eraser_path}.

\textbf{Temporal Reasoning.} Beyond reasoning over the current state, efficient path exploration requires an agent to reason over its past states. In CogExplore's case, we note the robot venturing to a new area despite having a point labeled as a fire extinguisher (the target exploration object). The robot already explored the nearby area in detail in a prior state, and the target object was never reached. The model correctly concludes the detection was in error. CogExplore justifies this change based on an understanding that object detections can be fallacious, which also highlights the framework's ability to perform fault-tolerant reasoning and remain robust to noisy sensor observations. In contrast, it was observed that the VEFEP planner would frequently stick around objects as the location of the 3D object projections were refined (as new observations were obtained), despite this not being an effective strategy after a few iterations.

We can also see from Figure \ref{fig:office_chair} that VEFEP suffered from exploring near the office chair but not actually reaching it. This is attributed to obstructed projections in the object detection system. Since the system performs a ray cast, obstructed views can cause object projections to fall short of the actual location. If no new observations update this position, then VEFEP will veer away from the area, whereas CogExplore will usually probe the area or come back to it if other cues indicate it's a good area to keep exploring.

\vspace{-8pt}
\section{Conclusions and Limitations}
Grounding vision models within simulated imagery poses challenges due to the limitations of VQA and object detection models, which can produce hallucinations, false negatives, and false positives. Our method addresses these errors by maintaining continuity in environment descriptions, limiting the number of target objects generated by the LLM, and requiring the LLM to interpret the environment across multiple generations. However, our approach lacks the ability to autonomously determine the correctness of identified target objects without ongoing experimentation and relies on simulation for confirmation or rejection, necessitating supervisor evaluation in potential deployments. Furthermore, while our discussion of results assumes the effectiveness of our approach based on its demonstrated performance and the compatibility of its justifications with the evaluated environments and objectives, we acknowledge the possibility of post hoc reasoning introducing a potential source of error.

CogExplore is a comprehensive framework for leveraging foundation models in robotic exploration tasks. The framework offers the ability to perform exploration that is geometrically, semantically, and  temporally aware, while remaining resilient to failures in grounding. Performance is evaluated across 210 photorealistic simulations in 3 different environments with 7 different exploration tasks. Our findings reveal that as the complexity of the exploration task increases, in terms of environment size and trajectory length, CogExplore's performance advantage over a vision-enabled exploration planner becomes more pronounced. This positive correlation shows that CogExplore is particularly well-suited for handling challenging navigation scenarios.

\label{sec:conclusion}


\acknowledgments{This work was supported by USDA-NIFA Award Number 2021-67021-3345}


\bibliography{main}  

\newpage
\section{Appendix}

\subsection{Overhead Path Views}

\begin{figure}[h!]
  \centering
  \includegraphics[width=0.6\textwidth]{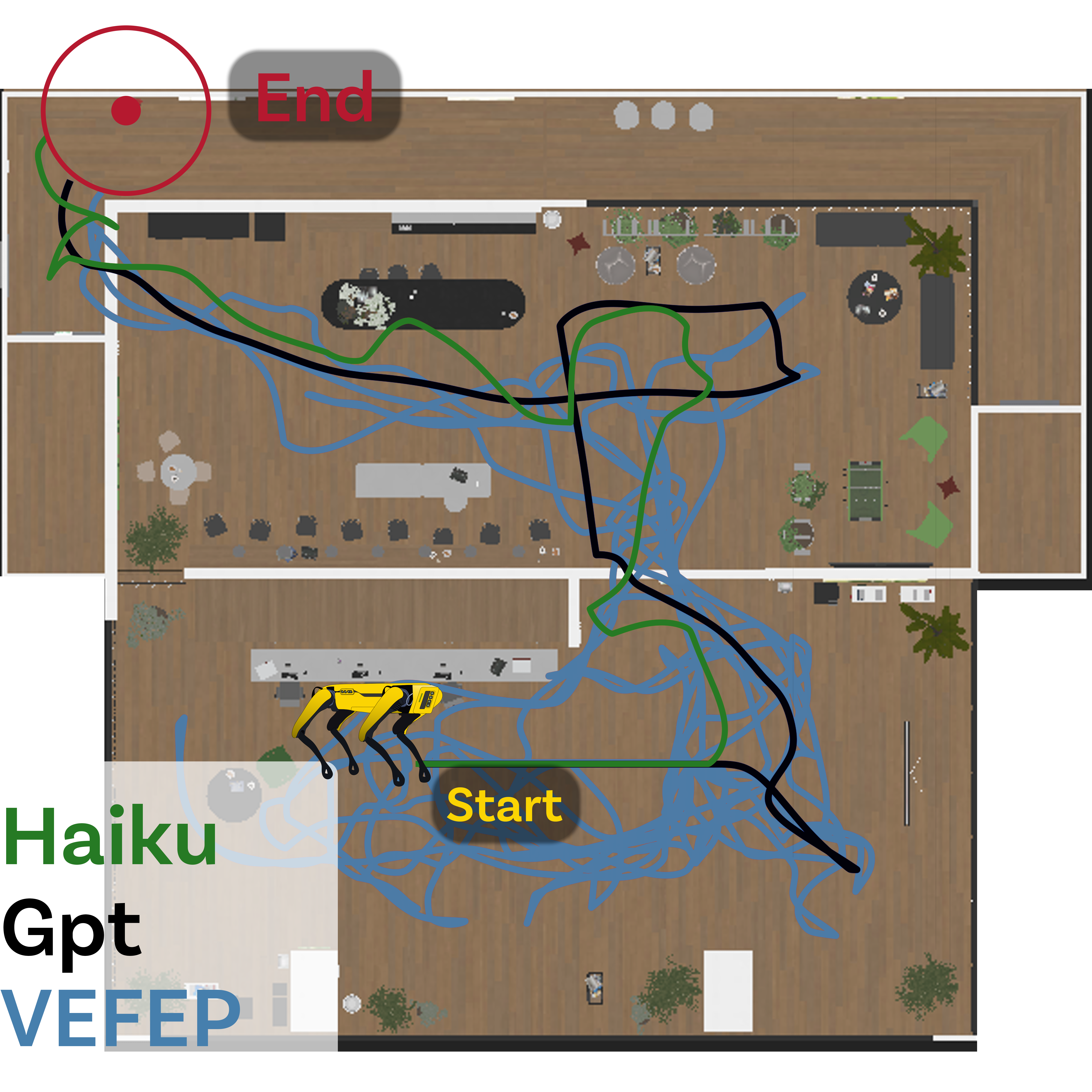} 
  \caption{Birds-eye view of Office Environment 1 (OE1) for task Fire Extinguisher 1 (FE1).}
  \label{fig:Overhead_OE1_FE1}
\end{figure}

\begin{figure}[h!]
  \centering
  \includegraphics[width=0.6\textwidth]{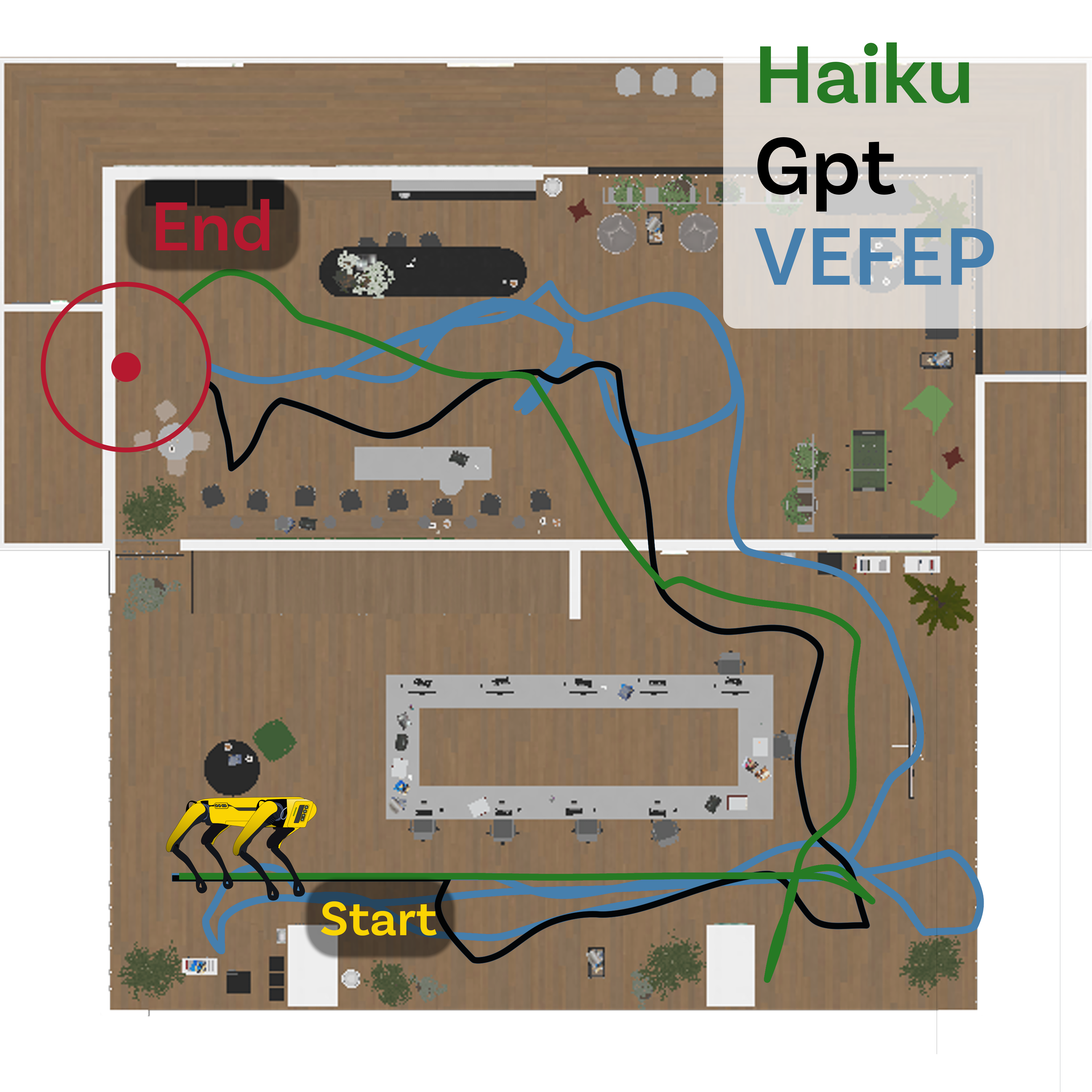}
  \caption{Birds-eye view of Office Environment 1 (OE1) for task Fire Extinguisher 2 (FE2).}
  \label{fig:Overhead_OE1_FE2}
\end{figure}

\begin{figure}[h!]
  \centering
  \includegraphics[width=0.6\textwidth]{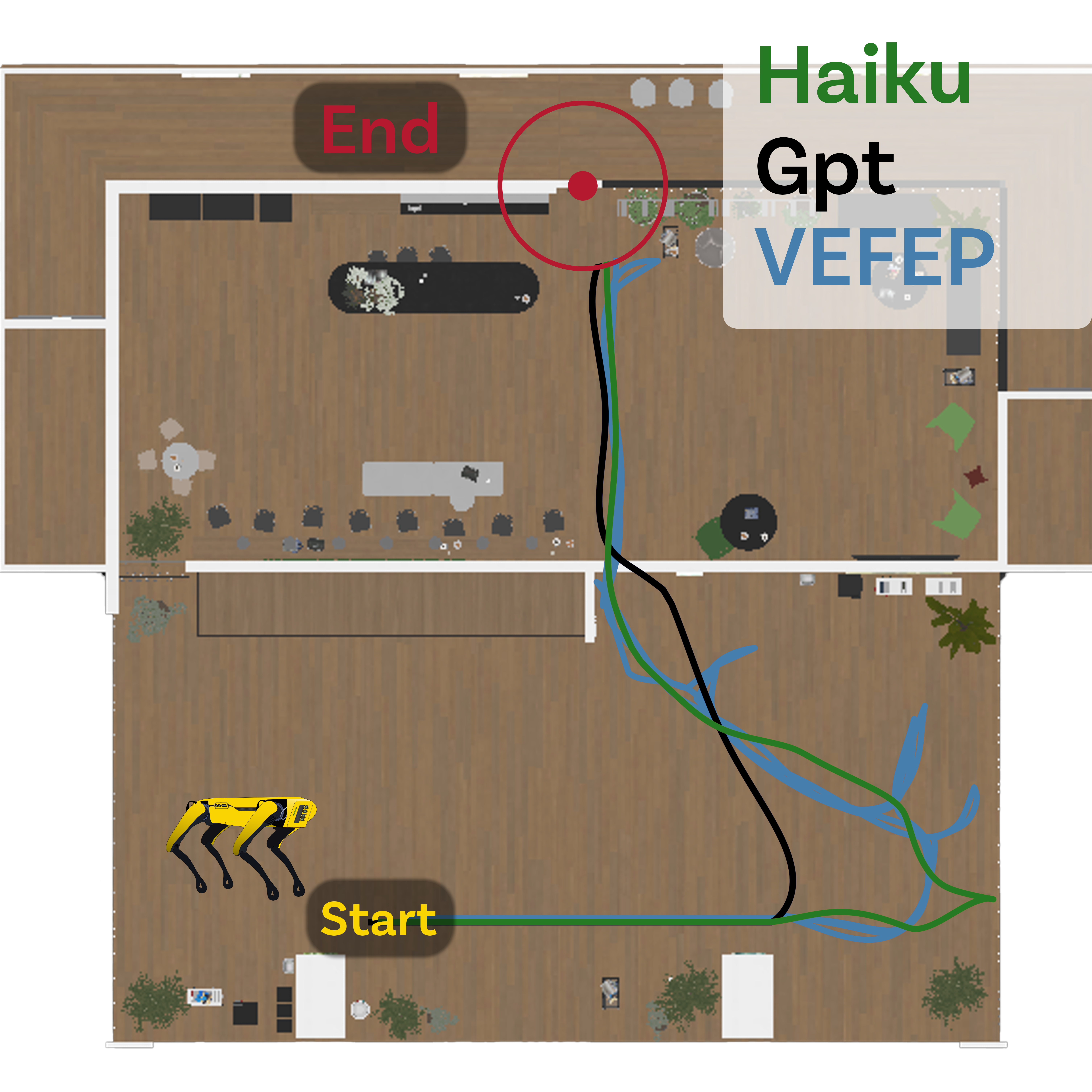}
  \caption{Birds-eye view of Office Environment 1 (OE1) for task Fire Extinguisher 1 (FE3).}
  \label{fig:Overhead_OE1_FE3}
\end{figure}

\begin{figure}[h!]
  \centering
  \includegraphics[width=0.6\textwidth]{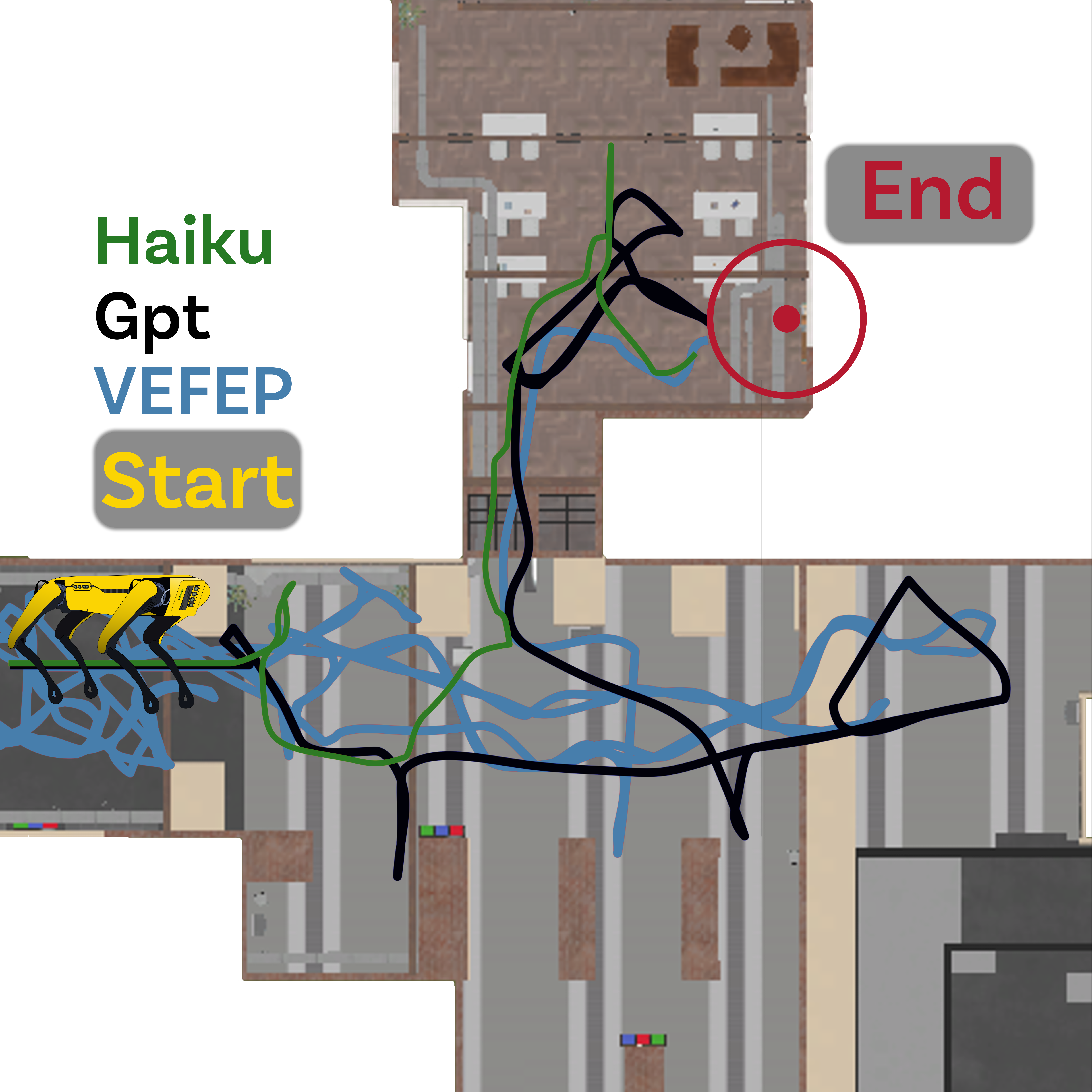}
  \caption{Birds-eye view of School for task Bookshelf}
  \label{fig:Overhead_Classroom}
\end{figure}

\begin{figure}[h!]
  \centering
  \includegraphics[width=0.6\textwidth]{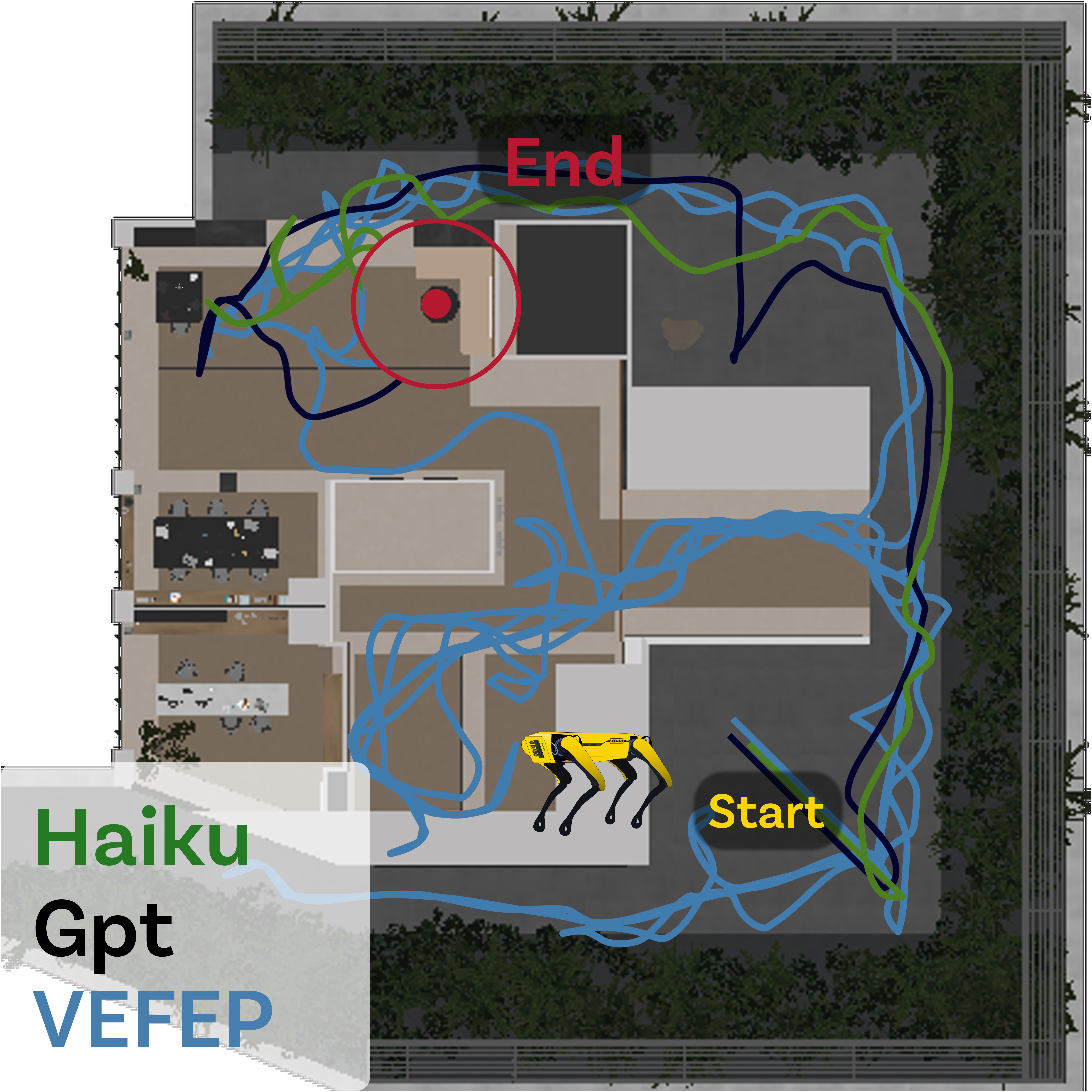}
  \caption{Birds-eye view of Office 2 for task Coffee Table}
  \label{fig:Overhead_OE2}
\end{figure}

\clearpage
\newpage

\subsection{Open Vocabulary Object Detection Pipeline}
\label{appendix:openvocab}

\begin{figure}
    \centering
    \includegraphics[width=0.9\textwidth]{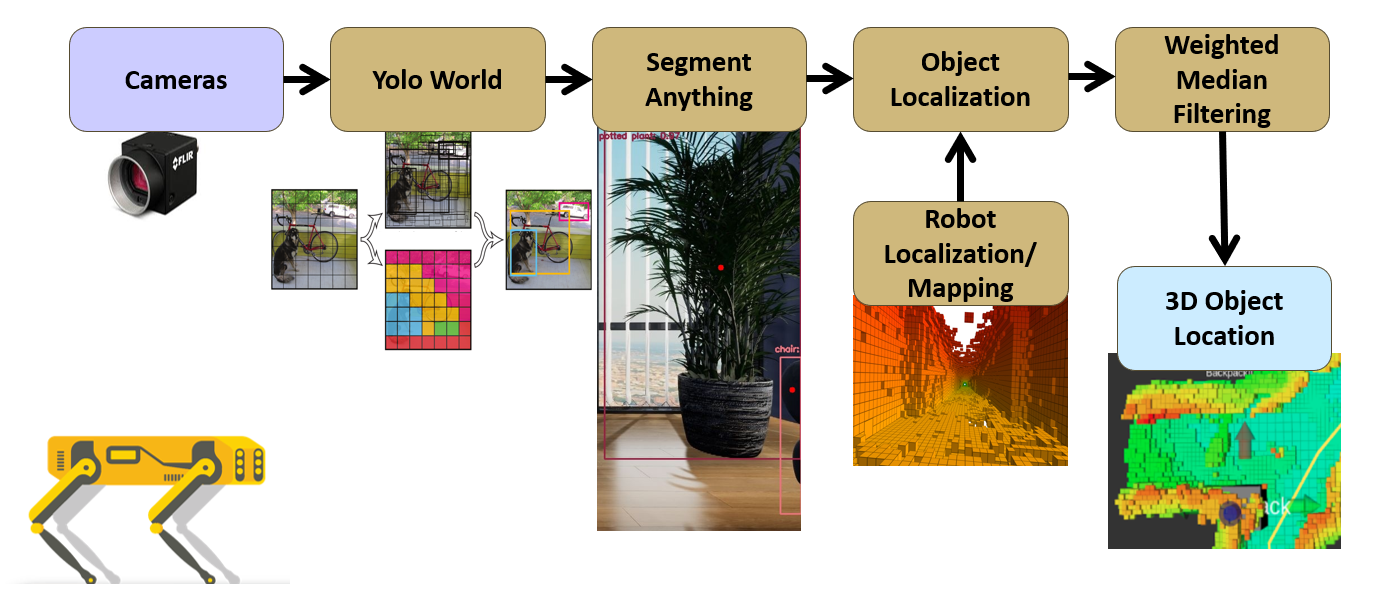}
    \caption{Open Vocabulary Detection Pipeline. The pipeline uses YOLO World \cite{Cheng2024YOLOWorld}, a variant of YOLO \cite{redmon2016you} to perform 2D detections which are then segmented by Segment Anything \cite{kirillov2023segment}. We take the centroid of the segmented object and perform a ray cast to project the object onto the robot's Octomap \cite{hornung2013octomap}. Projections are filtered using a weighted median filter. }
    \label{fig:objectpipeline}
\end{figure}

Our open vocabulary detection pipeline is shown in Figure \ref{fig:objectpipeline}.

\clearpage
\newpage

\subsection{Prompts For Foundation Models}
\label{appendix:prompting}

The full foundation model prompts for the foundation model are provided in the following Subsections. 

\subsubsection{VQA Prompting}
\label{appendix:vqa}

\begin{lstlisting}  
Your task is to create a list of 3 to 5 questions for a VQA (visual question answering model) that generate a clear and concise description of the enviornment that can later used by the robotic agent to navigate to the object or area described by this command: *INSERT_QUERY_HERE*
The description you elicit by providing questions to the VQA model is going to be passed to an operator who will use its content to make guesses about what objects in the environment to try to detect to assist with navigation.

VQA models are great at providing specific feedback to specific questions.

The robot you are guiding is the spot robot from Boston Dynamics.

The object or area we are after may be in plain view or may require exploration to discover. Do not presume the existence of any objects that you think may be in the scene.

For example, DO NOT ask any questions about the target area or object described in the command above. For example it is correct to ask if the target object is visible in the scene, but it is incorrect to ask for details about the target object since it may or may not be in the image.

Ask yes or no questions about whether an object exists or ask general questions about the kind of environment depicted within the image. DO NOT ASK ANY OTHER KIND OF QUESTION.

Your work will be added after two starter questions (do not include them in your response):
["What is the general setting of the environment: domestic, industrial, natural?", "What are the most prominent features visible in this environment?"]

Expand on these questions with 3 to 5 of your own. Do not replicate the work done by the above generic questions. Write your response formatted as a python list. That is a list denoted by brackets with each question separated by a comma. Only write this list. Do not include anything else in your response.

For example, a list with the two above questions included would look like: ["What is the general setting of the environment: domestic, industrial, natural?", "What are the most prominent features visible in this environment?"]

Note, your response must be a perfectly formatted list. You cannot include any other comments or characters in your response that are not precisely formatted in the same manner as the list example above. Make sure to include commas in between entries within the list. Make sure to include quotes around each entry within the list. These are strings within python, they require quotes.

Do not include any of the starter questions in your response. Just include the questions you have thought up.

The VQA model does not know anything about the position of the robot or that it is examining images from a robot. The questions must be straightforward and directly about the image itself.

\end{lstlisting}

\subsubsection{Object Detection Label Prompting}
\label{appendix:objectprompting}

\begin{lstlisting}
*INSERT_SCENE_DESCRIPTION*

Given you have this description of an environment we want to identify a set of objects that help us navigate a robotic agent to fulfill the goal: *INSERT_QUERY_HERE*

Given this information, generate a comprehensive list of all objects including the objects referenced in the prompt that may help the robotic agent fulfill the navigation request. This list will be used to guide an object detection machine learning model. So words in this list should be general and likely to be spotted within 
the environment as we have characterized it in the scene description details above. You should look for specific individual items, such as tables, and plants that can be easily distinguished from the background. DO NOT look for general entities that could take up the entire frame such as wall, floor, path, road, or park. Lables like this should not be in your list.
Do not look for very large objects.

Always include the object or area you are searching for in this list. Rember this object detector is running in realtime on a moving robot.

This list should be between 1 and 5 objects in length. It should be formatted as a python list. For example, a list of the following objects would be precisely formatted as: ["chair", "table", "helmet", "tree"]
Note the first entry must be the goal object!

This list is the only thing that should be included in your response. The python list format must be followed precisely. Be sure to include quotes around the elements of the list.

Do not return anything besides this list.

\end{lstlisting}

\subsubsection{Explore Iteration Prompting}
\label{appendix:explore prompting}

\begin{lstlisting}
    You are guiding a robotic agent to solve the following robotic navigation task *INSERT_QUERY_HERE*. 
You are assisted by an exploration planner that produces potential new areas to explore in the form of graph and frontier points. Frontier points tend to be in areas you have explored less or have just discovered.
There is also an onboard object detector that produces 3D projections of points in space that correspond to objects from a given object list. Remember object points represent the centroid for an object and your planner will get you as close as it can to them.
Object points are often not perfectly labeled. Frontier points are calculated based on exploration potential. Object points will come from areas you have already observed.
Your objective is to select a point from among these within the scene, and determine what will serve as the robot's next waypoint.
You will only have waypoints in areas your robot can see with its lidar and cameras. So, it is important to probe areas that potentially offer vantages for views you have not yet seen. When you see new areas, you should explore them.
If you have been near a target for more than one call to this point selection logic, then it is not the correct target. The computer vision your robot has is not perfect but is tuned towards producing more false postives. Be adventurous in your exploration, travel to areas far way from you. 
Remember your robot is continuously detecting objects and has a left, right and front cameras. When you select a point the planner will plan a direct route for your robot to take. If the target object is deteceted along your path, the robot will go to it and add an entry to the summary below. 
Use the available information from the robot below along with any contextual information you have available to select the next point to navigate to.

The time since start of sim is:
*INSERT_CURRENT_TIME* seconds

The spot robot is now at:
*INSERT_CURRENT_POSITION*

Here is a description of the area spot is currently in:
*INSERT_SCENE_DESCRIPTION*

The robot has been instructed to search for these objects:
*INSERT_OBJECT_LIST*

You have the following list of points you may choose from (graph points are not labeled, frontier points and object points are). If a graph point is labeled as "new" then it is from a region you have discovered it in the last step and are now able to plan to.
You should go to new points! They represent unknown areas and new views for your cameras and lidar. You may select either a graph point or an object point.
*INSERT_FRONTIER_OBJECT_POINT_NUMBERED_LIST*

You have been asked to help in the past *INSERT_TOTAL_CALLS* times to inform this navigation task. You should never go to the same point multiple times. The simulation ends when you reach the target.

Here is a summary of the actions you performed in the last *INSERT_MEMORY_LENGTH* calls:
*INSERT_PRIOR_STATES*

*INSERT_INTERRUPT_DESCRIPTION*


Now that you know where you've been, try to explore areas you have not yet been to. Use the description of the environment in past states to go where the object is most likely. If you have no prior states, be aggressive in your exploration.

If you have been by a point for more than one call, you should move, even if it is the target object. The robot's vision makes mistakes. Do not go to the same point more than once! Check the coordinates (x,y,z) of the point, if you've gone near there before in any of these prior states, don't go there again. 
Change points every time, even if the point is an object point which appears to be the target. Cycling back and forth between two areas will not give new observations. The environment is static if you have already been to an area your cameras should have found the object.

As you explore the scene, new points will become available for you to plan to. Try to select points on the edge of those available to you. Consider areas that will allow you to discover new points. If a point is far from the other points in your list and in your prior states, it is a good candidate.

Reason about the area of space defined by your points. Go to areas and rooms outside of spaces you have explored.

Avoid considering points near your current location if you do not believe they are near the target.

Do not worry about collisions. All points are safe to travel to. Explore the space aggressively. Seek out the object and trust the robot's vision to find it.

If you have a point or points which are labeled as a new point, you should go to one of these points, your task is to to explore with contextual awareness. These new points are newly discovered and represent a high potential for discovering your target.

Recall, you should consider x, y, and z when deciding where to go for both the object locations and the potential frontier points you want to go to. Do not just look at one coordinate! Consider both x and y equally! Euclidian distance should be used, we want to be close to the target object across all dimensions.
The distance field in your point list is the Euclidian distance from the robot to the object.

NEVER GO TO THE SAME AREA TWICE IN A ROW!!!

*INSERT_SPECIFIC_MODEL_INSTRUCTIONS*

Based on the information above fill out the form below (if you select an object, make sure to explain why that object is related to the navigation task. Reference past states in your reasoning. You should always navigate in reference to where you've been. Reason about the space you are in. Reference the objects you see and categorize your environment.):

I need to select a waypoint from a numbered list of graph points, frontier points and object points. The point I have selected is point number: [insert the graph, frontier or object point numeral number here]. I am selecting this point because I believe it makes strategic sense to get me closer to solving my navigation task "*INSERT_QUERY_HERE*". My environment can be described as [describe the environment]. My reasoning is that this point [insert your reasoning here].
\end{lstlisting}

\subsubsection{GPT-3.5 Turbo Specific Prompt Additions}

\begin{lstlisting}
If you see the object go directly to it.

Avoid going to regions that you have been to before. Consider how close the point you are selecting is with those listed in your prior states. If it is nearby, don't return to it.

Use your vision to go to areas that are rich with objects like the one you are after.

Newly discovered areas that are far away are highly advantageous if they are far from where your robot has traveled before.

Use the points to gain a sense of space. Travel to regions where you have not yet been.

Don't be tricked by windows, if you can't get to interesting objects, they might be behind a window.

GO TO AREAS THAT ARE FAR AWAY! DON'T STAY IN ONE REGION!!
\end{lstlisting}

\subsubsection{Claude Haiku Specific Prompt Additions}

\begin{lstlisting}
Avoid looping between areas you have already been to. Try to find new areas so you will generate graph points for them. If there is an area you have not been to yet, go to it. Never double back, keep exploring nearby unknown areas.

Consider each prior state, do not return to anywhere near these locations.

Newly discovered areas that are far away are highly advantageous if they are far from where your robot has traveled before.

Use the points to gain a sense of space. Travel to regions where you have not yet been.

Always prioritize new points when they are available.

GO TO AREAS THAT ARE FAR AWAY! DON'T STAY IN ONE REGION!!
\end{lstlisting}

\end{document}